\documentclass[10pt,twocolumn,letterpaper]{article}

\usepackage{iccv}
\usepackage{amsmath,amssymb} 
\usepackage{graphicx}
\usepackage{times}
\usepackage{epsfig}
\usepackage{color}
\usepackage{algorithm}
\usepackage{algorithmic}
\usepackage{multirow, makecell}
\usepackage[table,xcdraw]{xcolor}
\usepackage[labelsep=period]{caption}
\usepackage{subcaption}
\usepackage{url}

\usepackage[pagebackref=true,breaklinks=true,letterpaper=true,colorlinks,bookmarks=false]{hyperref}

\iccvfinalcopy 

\def\iccvPaperID{872} 

\ificcvfinal\fi
\begin{document}

\title{Exploiting Multi-layer Graph Factorization for \\Multi-attributed Graph Matching}

\author{Han-Mu~Park\\
	Gwangju Institute of Science and Technology\\
	{\tt\small hanmu@gist.ac.kr}
	\and
	Kuk-Jin~Yoon\\
	Gwangju Institute of Science and Technology\\
	{\tt\small kjyoon@gist.ac.kr}
}

\maketitle

\begin{abstract}
	\vspace{-7pt}
Multi-attributed graph matching is a problem of finding correspondences between two sets of data while considering their complex properties described in multiple attributes.
However, the information of multiple attributes is likely to be oversimplified during a process that makes an integrated attribute, and this degrades the matching accuracy.  
For that reason, a multi-layer graph structure-based algorithm has been proposed recently.
It can effectively avoid the problem by separating attributes into multiple layers. 
Nonetheless, there are several remaining issues such as a scalability problem caused by the huge matrix to describe the multi-layer structure and a back-projection problem caused by the continuous relaxation of the quadratic assignment problem. 
In this work, we propose a novel multi-attributed graph matching algorithm based on the multi-layer graph factorization.
We reformulate the problem to be solved with several small matrices that are obtained by factorizing the multi-layer structure.
Then, we solve the problem using a convex-concave relaxation procedure for the multi-layer structure.
The proposed algorithm exhibits better performance than state-of-the-art algorithms based on the single-layer structure.
\end{abstract}

\vspace{-9pt}
\section{Introduction}
\label{sec_introduction}

Graph matching is a problem of finding correspondences between two sets of data while considering their structural information.
Since the structured data is robust against deformation and outliers, graph matching has been widely adopted to formulate various correspondence problems in computer vision such as tracking,  detection~\cite{chen2001multi,gomila2003graph,xiao2010vehicle}, shape matching~\cite{duchenne2011tensor,huang2012optimization,leordeanu2005spectral}, and object classification~\cite{cho2013learning,duchenne2011graph}.
Essentially, a graph structure consists of a set of vertices and a set of edges.
However, in recent studies, attributes describing characteristics of vertices and edges are also commonly included in the structure for practical reasons such as matching accuracy or computational complexity~\cite{conte2004thirty,yan2016short}.
Since the attributes provide important information for the vertices and edges, how to define and how to utilize the attributes significantly affects the matching performance.

There are two types of attributes that are commonly employed in computer vision problems: appearance and geometric.
The former describes the appearance of an element that should be matched such as color and shape~\cite{yan2015consistency,zhou2012factorized}, which is often used as the vertex attribute.
The latter describes the geometric relationships between elements such as angles and distances~\cite{cho2013learning,cho2010reweighted,cour2006balanced,leordeanu2005spectral,yan2015consistency,zhou2012factorized}, which is commonly used as the edge attribute.
Since a single-type attribute does not have enough capability to describe complicated properties of image contents in general, many previous works~\cite{cho2013learning,cho2010reweighted,cour2006balanced,duchenne2011tensor,gomila2003graph,leordeanu2005spectral,zhou2012factorized} adopted multiple attributes and constructed a mixed-type attribute by integrating multiple single-type attributes. 
%
%
%
In this case, although the integrated attribute can provide better representation for real-world applications than the single-type attribute, there are a few fundamental problems caused by the properties of integration approaches as stated in \cite{park2016multi}.
First, the distinctive and rich information from multiple attributes can be distorted or lost during the integration process.
Because the attributes have different properties from each other, the characteristics of one attribute can be oversimplified by the other attributes as a consequence of the attributes integration. 
Second, the combination of multiple attributes cannot be adaptively modified or changed once it is determined. 
Since each application has different characteristics and different setting, designing a versatile integration method that is suitable for all applications is impractical. 
Unfortunately, because integrated attributes cannot be decomposed and re-integrated, the over-fitting problem caused by the inflexible attribute combination arises.

\begin{figure*}[t!]
	\centering
	\includegraphics[width=0.83\linewidth]{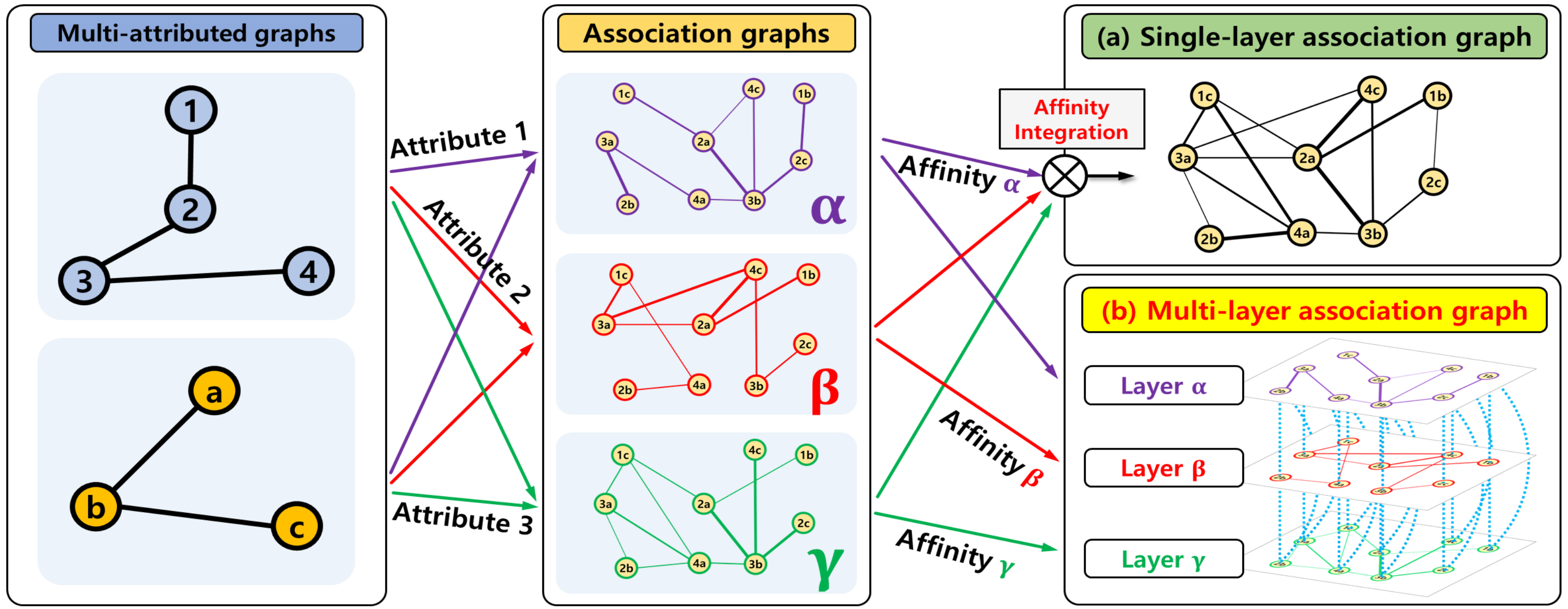}
	
	\vspace{-5pt}	
	\caption{Two types of problem formulation for graph matching with multiple attributes.
		(a)~An integrated affinity matrix is constructed by combining multiple affinity matrices of multiple attributes.
		(b)~Each of multiple attributes is separately represented, and affinity matrices are constructed for multiple layers that are linked to one another.
	} 
	
	\label{fig_MainConcept} \vspace{-3pt}
\end{figure*}

Recently, Park and Yoon~\cite{park2016multi} tried to solve the problems of integration approaches by adopting a multi-layer structure.
They proposed a multi-layer structure to consider multiple attributes jointly while preserving the properties of the attributes.
In the structure, each attribute is described in each layer, and the layers are connected to each other as shown in Fig.~\ref{fig_MainConcept}.
Thanks to the separation of the layers, the distinctive information of each attribute can be preserved.
Furthermore, the relationships between the attributes can be redefined during the matching process by manipulating inter-layer links. 
Based on the structure, they proposed a random walks based graph matching algorithm for the multi-layer structure, which is called multi-layer random walk matching~(MLRWM).
The algorithm provides a more robust performance than state-of-the-art single-layer graph matching algorithms against outlier attributes. 
However, since the multi-layer structure requires a huge matrix whose dimensions are proportional to the number of attributes, this can cause a scalability problem.
Moreover, because the objective function of MLRWM is still non-convex as in the single-layer random walk matching algorithm~\cite{cho2010reweighted}, it is easy for the algorithm to fall into a local optimum.

In this paper, we propose a multi-layer graph matching algorithm based on the multi-layer graph factorization to solve multi-attributed graph matching problem.
The contributions of this paper are threefold.
First, we propose a graph factorization method for the multi-layer structure considering the block structure of a supra-adjacency matrix. 
Second, by using the results of the factorization, we formulate a multi-attributed graph matching problem with the convex and concave relaxation.
Finally, we propose a multi-layer path following algorithm based on the relaxation formulations by generalizing the single-layer algorithm~\cite{zhou2012factorized,zhou2013deformable}.

\section{Related Works}
\label{sec_related_works}
\vspace{-1pt}

Recently, the convex-concave relaxation-based graph matching algorithms~\cite{liu2014gnccp,liu2012extended,liu2014graph,zaslavskiy2009path,zhou2012factorized,zhou2013deformable} have received wide attentions because of their superior performance.
The algorithms based on the so called path following approach has two relaxed functions in common: a convex function and a concave function.
The convex function has an approximated optimum, which is different from that of the original objective function, but the function converges very quickly in exchange.
By contrast, the global optimum of the concave function is exactly the same as that of the original objective function; however, it is difficult to find the global optimum because there are multiple local optima.
The objective function of the path following scheme is constructed by combining two relaxed functions.
Then, the function gradually moves an approximated optimum of the convex function to a discrete optimum of the concave function.
Algorithms that follow the spectral approximation strategy~\cite{cho2013learning,cho2010reweighted,cour2006balanced,leordeanu2005spectral} solve the problem in the continuous domain, and then reproject the solution into a discrete domain by using a discretization method such as the Hungarian algorithm~\cite{munkres1957algorithms}.
On the other hand, the path following strategy can generate a discretized solution after the optimization process.
This can allow to avoid additional errors caused by the discretization process, as stated in \cite{zhou2012factorized}.

The path following algorithm was firstly proposed by Zaslavskiy~\etal~\cite{zaslavskiy2009path}. 
They formulated a graph matching problem on an adjacency matrix, and solved the problem by using a convex-concave relaxation approach. 
Zhou and De la Torre~\cite{zhou2012factorized,zhou2013deformable} proposed the methods that factorize the affinity matrix, and applied the algorithm to the graph matching. 
In their work, the convex-concave relaxation of the objective function was derived by using the factorized matrices. 
Liu~\etal~\cite{liu2012extended} extended the original algorithm for solving the directed graph matching problems defined on an adjacency matrix by modifying the concave relaxation method.
They then generalized the algorithm to solve the problems that are defined for the affinity matrix and partial permutation matrix respectively in~\cite{liu2014gnccp,liu2014graph}.

The proposed algorithm is inspired by the factorization-based convex-concave relaxation approach~\cite{zhou2012factorized,zhou2013deformable}.
Since multi-attributed graph matching problems that are formulated using a multi-layer structure require a large matrix to describe its structure, the factorization-based relaxation scheme is very useful in resolving the scalability issue.

\section{Problem Formulation}
\label{sec_problem_formulation}
\vspace{-2pt}

Attributed graph matching problems can be categorized into two types according to the description style of attributes: single-layer structure graph matching and multi-layer structure graph matching. 

\begin{figure*}[t!]
	\centering
	\includegraphics[width=0.99\linewidth]{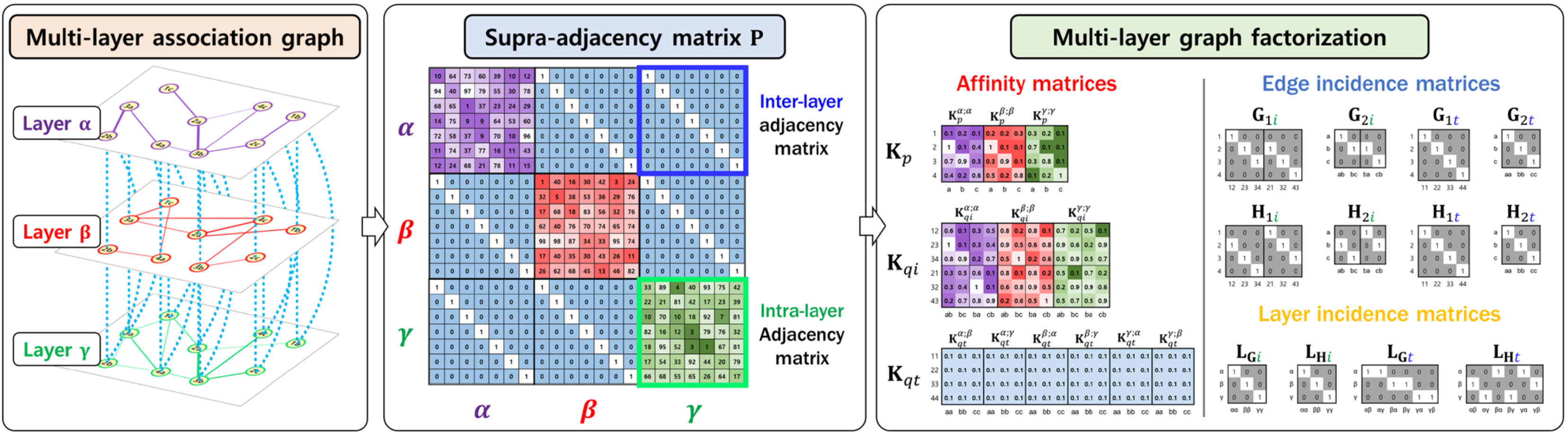} 
	\vspace{-6pt}
	\caption{Supra-adjacency matrix $\mathbf{P}$ that describes the multi-layer structure can be factorized into a series of small matrices.} 
	\label{fig_algorithm_concept} \vspace{-6pt}
\end{figure*}

\subsection{Single-layer graph matching problem}
\label{sec_problem_formulation_single}

A single-layer graph matching problem finds correspondences between two attributed graphs that are described as single-layer structures.
This includes not only single-attributed graph matching but also multi-attributed graph matching problems, because multiple attributes can be considered as a single attribute by integrating them.

For given two attributed graphs $\mathcal{G}_{1}$ and $\mathcal{G}_{2}$, this problem is commonly formulated as a Lawler's quadratic assignment problem (QAP)~\cite{cho2010reweighted,cour2006balanced,leordeanu2005spectral,park2016multi,zhou2012factorized}.
Each graph $\mathcal{G}_{k}$ is defined as a set $\left\{\mathcal{V}_{k},\mathcal{E}_{k},\mathcal{A}_{k}\right\}$, where $\mathcal{V}_{k}$ is a set of vertices, $\mathcal{E}_{k}$ represents a set of edges, and $\mathcal{A}_{k}$ is a set of attributes that describe each vertex and edge.
In the Lawler's QAP formulation, the correspondences between two sets of vertices are represented using an $({N}_{1}\times{N}_{2})$-dimensional binary assignment matrix $\mathbf{X}$, where ${N}_{1}$ and ${N}_{2}$ represent the numbers of vertices in the graphs.
Each element of the matrix $\left[\mathbf{X}\right]_{i,a}$ indicates a correspondence relation between ${v}_{(1)i}\in\mathcal{V}_{1}$ and ${v}_{(2)a}\in\mathcal{V}_{2}$.
For example, if ${v}_{(1)i}$ and ${v}_{(2)a}$ are matched, $\left[\mathbf{X}\right]_{i,a}=1$, otherwise $\left[\mathbf{X}\right]_{i,a}=0$.
On the other hand, the affinity information between two matching candidates is described using an $({N}_{1}{N}_{2}\times{N}_{1}{N}_{2})$-dimensional matrix $\mathbf{K}$.
Each diagonal element of the matrix $\left[\mathbf{K}\right]_{ia,ia}$ describes a unary affinity of a matching candidate $\left({v}_{(1)i},{v}_{(2)a}\right)$, and each non-diagonal element $\left[\mathbf{K}\right]_{ia,jb}$ describes a pairwise affinity of two matching candidates $\left({v}_{(1)i},{v}_{(2)a}\right)$ and $\left({v}_{(1)j},{v}_{(2)b}\right)$.
These affinity values are computed by using the vertex and edge attributes.
Then, the graph matching problem can be formulated as follows:
\begin{equation}
\hat { \mathbf{X} } = \underset { \mathbf{X} } {arg\max } {\left( {  \text{vec}\left(\mathbf{X}\right)  }^{ \top } \mathbf{K}  \text{vec}\left(\mathbf{X}\right)  \right)}, 
\label{eq_problem_single_layer_original}
\end{equation}
\begin{equation*}
s.t.\quad \mathbf{X} \in { \left\{ 0,1 \right\}  }^{ { N }_{ 1 }\times{ N }_{ 2 } }, \quad \mathbf{X}\mathbf{1}_{{N}_{2}}\le\mathbf{1}_{{N}_{1}}, \mathbf{X}^{\top}\mathbf{1}_{{N}_{1}}\le\mathbf{1}_{{N}_{2}}, 
\end{equation*}
where $\mathbf{1}_{N}$ is an $N$-dimensional all-ones vector, and the vector inequality constraints represent the one-to-one matching constraint of the graph matching problem.
Since this QAP is a well-known NP-hard problem, previous works~\cite{cho2010reweighted,cour2006balanced,leordeanu2005spectral,park2016multi,zhou2012factorized} addressed the problem by relaxing the binary matching constraint to find an approximated solution in the continuous domain.

\subsection{Multi-layer graph matching problem}
\label{sec_problem_formulation_multi}

The formulation of multi-layer graph matching was firstly proposed in \cite{park2016multi} to solve multi-attributed graph matching problems.
They proposed a multi-layer association graph structure inspired by the multiplex network structure~\cite{de2013centrality,kivela2014multilayer,sole2015random}. 
The graph $\mathcal{G}_{k}\left(\mathcal{V}_{k},\mathcal{E}_{k},\mathcal{A}_{k},\mathcal{L}_{k}\right)$ consists of multiple layers that describe multiple attributes respectively, and the layers are linked to each other as shown in Fig.~\ref{fig_algorithm_concept}. Here, $\mathcal{L}_{k}$ is a set of layer indices. 
In contrast to the single-layer structure, each vertex ${v}^{\alpha}_{(k)i}\in\mathcal{V}_{k}$ has a layer index $\alpha\in\mathcal{L}_{k}$.
For this reason, vertices that have different layer indices, ${v}^{\alpha}_{(k)i}$ and ${v}^{\beta}_{(k)i}$, are differentiated from each other, even though they have the same vertex index.
Accordingly, each edge ${e}^{\alpha;\beta}_{(k)i;j}\in\mathcal{E}_{k}$ is also defined with two layer indices.
Thus, a four-dimensional affinity tensor $\mathbf{\Pi}$ is required to express the multi-layer association graph structure as described in \cite{park2016multi}.
$\mathbf{\Pi}$ consists of two types of affinity matrices (intra- and inter-layer), which are arranged as a two-dimensional structure according to layer indices.
$\mathbf{\Pi}$ can be flattened to a two-dimensional block matrix as shown in Fig.~\ref{fig_algorithm_concept}, which is called the supra-adjacency matrix~\cite{de2013centrality,gomez2013diffusion,park2016multi,sole2015random}.
Then, a multi-layer graph matching problem can be formulated as 
\begin{equation}
\begin{split}
\hat{\textbf{X}} &= \underset { \textbf{X} } {arg\max } \quad {F}_{gm}\left(\textbf{X}\right) \\
&= \underset { \textbf{X} } {arg\max } \left(\textbf{L}_{\mathbf{C}}\otimes\text{vec}\left(\textbf{X}\right)\right)^{\top}\textbf{P} \left(\textbf{L}_{\mathbf{C}}\otimes\text{vec}\left(\textbf{X}\right)\right) \\
s.t. & \mathbf{X} \in { \left\{ 0,1 \right\}  }^{ { N }_{ 1 }\times{ N }_{ 2 } },  \mathbf{X}\mathbf{1}_{{N}_{2}}\le\mathbf{1}_{{N}_{1}}, \mathbf{X}^{\top}\mathbf{1}_{{N}_{1}}\le\mathbf{1}_{{N}_{2}},
\end{split}
\label{eq_multi_attributed_graph_matching_problem}
\end{equation}
where $\otimes$ indicates the Kronecker product, and $\mathbf{P}$ is a supra-adjacency matrix.
$\mathbf{L}_{\mathbf{C}}$ is an ${N}_{L}$-dimensional vector that describes relative confidence values among attributes, and ${N}_{L}$ represents the number of layers.
Details of the $\mathbf{L}_{\mathbf{C}}$ estimation are presented in Sec.~\ref{sec_relative_confidence_computation}.
Although this multi-layer based formulation effectively resolves the oversimplification problem by preserving characteristics of each attribute, the scalability problem caused by the huge size of the supra-adjacency matrix comes to the fore in exchange.
To solve this scalability problem, we propose a matrix factorization method that divides the huge multi-layer matrix into several small matrices.
Details are presented in Sec.~\ref{sec_multi_layer_graph_factorization}.

\section{Multi-layer Graph Factorization}
\label{sec_multi_layer_graph_factorization}

The supra-adjacency matrix consists of two types of block matrices (intra- and inter-layer affinity matrices), and these matrices have totally different characteristics as shown in Fig.~\ref{fig_algorithm_concept}.
Zhou and De~la~Torre~\cite{zhou2013deformable} proposed a factorization method for the single-layer graph matching problem. They divided the affinity matrix into unary and pairwise affinity matrices and several edge incidence matrices.
Similarly, we categorize the relations between matching candidates into three types of affinity matrices: unary, pairwise intra-layer, and pairwise inter-layer matrices.
Then, we represent the supra-adjacency matrix by combining the affinity matrices and incidence matrices.

Specifications of the matrices are presented in Table~\ref{tab_matrix_specification} and Fig.~\ref{fig_algorithm_concept}.
In contrast to the single layer structure, since each vertex of the multi-layer structure has intra- and inter-layer relations, the edge incidence matrix and layer incidence matrix should be used simultaneously to indicate a specific relation.
$\mathbf{G}$ and $\mathbf{H}$ are edge incidence matrices, and each column of the matrices represents starting or ending vertex of each edge.
To represent directional edges, edges are differentiated from each other according to their starting vertices.
The subscripts of $\mathbf{G}$ and $\mathbf{H}$ represent the types of affinity matrices.
For example, the subscript $1i$ indicates intra-layer connections of $\mathcal{G}_{1}$, and the subscript $2t$ indicates inter-layer links of $\mathcal{G}_{2}$.
$\mathbf{L}_{\mathbf{G}}$ and $\mathbf{L}_{\mathbf{H}}$ are layer incidence matrices, and the matrices describe the relation between layers similar with $\mathbf{G}$ and $\mathbf{H}$.
On the other hand, the affinity information of the multi-layer structure is categorized into three types as mentioned above.
The unary affinity matrix $\mathbf{K}_{p}$ is constructed by concatenating layerwise unary affinity matrices such as $\mathbf{K}^{\alpha;\alpha}_{p}$.
$\mathbf{K}_{qi}$ and $\mathbf{K}_{qt}$ are pairwise intra-layer and inter-layer affinity matrices, which are also constructed by concatenating layerwise affinity matrices.
Finally, the supra-adjacency matrix can be represented by combining these incidence matrices and affinity matrices as follows:

\begin{table}[t!]
	\small
	\centering
	\caption{Matrix specification} \vspace{-6pt}
	\label{tab_matrix_specification}
	{
    \renewcommand{\tabcolsep}{1.5mm}
		\begin{tabular}{m{3.5cm}|m{4.4cm}}
			\noalign{\hrule height 0.5pt} \hline			
			Matrix      &  Specification                                    \\ \hline		\hline
			$\textbf{G}_{1i}$, $\textbf{G}_{2i}$, $\textbf{G}_{1t}$, $\textbf{G}_{2t}$	& 	Edge incidence matrix (start)\\
			$\textbf{H}_{1i}$, $\textbf{H}_{2i}$, $\textbf{H}_{1t}$, $\textbf{H}_{2t}$	&	Edge incidence matrix (end)\\
			$\textbf{L}_{\textbf{G}i}$, $\textbf{L}_{\textbf{G}t}$						&	Layer incidence matrix (start)\\ 
			$\textbf{L}_{\textbf{H}i}$, $\textbf{L}_{\textbf{H}t}$  					&	Layer incidence matrix (end)\\
			${\textbf{K}_{p}=}  \left[\textbf{K}_{p}^{\alpha;\alpha}\textbf{K}_{p}^{\beta;\beta}\cdots\textbf{K}_{p}^{\omega;\omega}\right]$ & Unary affinity matrix	\\ 
			${\textbf{K}_{qi}=}  \left[\textbf{K}_{qi}^{\alpha;\alpha}\textbf{K}_{qi}^{\beta;\beta}\cdots\textbf{K}_{qi}^{\omega;\omega}\right]$	&  Pairwise intra-layer affinity matrix\\ 
			${\textbf{K}_{qt}=} \left[\textbf{K}_{qt}^{\alpha;\beta}\textbf{K}_{qt}^{\alpha;\gamma}\cdots\textbf{K}_{qt}^{\psi;\omega}\right]$	&  Pairwise inter-layer affinity matrix \\ 
			\noalign{\hrule height 0.5pt} \hline	
		\end{tabular}
	} 
\end{table}

\begin{equation}
\small
	\begin{split}
	&\textbf{P} = \text{diag}\left(\text{vec}\left(\textbf{K}_{p}\right)\right) \\ & + \left(\textbf{L}_{\textbf{G}i}\otimes\textbf{G}_{2i}\otimes\textbf{G}_{1i}\right)\text{diag}\left(\text{vec}\left(\textbf{K}_{qi}\right)\right)\left(\textbf{L}_{\textbf{H}i}\otimes\textbf{H}_{2i}\otimes\textbf{H}_{1i}\right)^{\top}\\
	& + \left(\textbf{L}_{\textbf{G}t}\otimes\textbf{G}_{2t}\otimes\textbf{G}_{1t}\right)\text{diag}\left(\text{vec}\left(\textbf{K}_{qt}\right)\right)\left(\textbf{L}_{\textbf{H}t}\otimes\textbf{H}_{2t}\otimes\textbf{H}_{1t}\right)^{\top},
	\end{split}
	\label{eq_supra_adjacency_matrix_factorization}
\end{equation}
where the first term indicates the unary affinity information, and the second and third terms represent the pairwise affinity information.

This factorized formulation has two advantages.
First, by dividing a huge matrix into several small matrices, a more efficient representation than the original matrix is available.
Second, by using the factorized formulation, 
the multi-layer graph matching problem can be relaxed into convex and concave functions.
This means that the path following approach~\cite{zaslavskiy2009path,zhou2012factorized,zhou2013deformable} can be applied to optimize the relaxed objective function, which exhibits a state-of-the-art performance.


\section{Multi-layer Factorized Graph Matching}
\label{sec_multi_layer_factorized_graph_matching}

In this section, we propose a multi-layer factorized graph matching algorithm by generalizing the path following algorithm~\cite{zaslavskiy2009path,zhou2012factorized,zhou2013deformable}.
To applying the path following algorithm, convex and concave relaxations of the objective function are required.
Thus, we firstly propose two relaxations for the factorized supra-adjacency matrix, and then propose a multi-layer path following algorithm.\footnote{Detailed  derivations and proofs of equations are presented in our supplementary materials.}

\subsection{Convex-concave relaxation}
\label{sec_convex_concave_relaxation}


Before relaxing the objective function ${F}_{gm}\left(\mathbf{X}\right)$, we derive an equivalent objective function by substituting Eq.~(\ref{eq_supra_adjacency_matrix_factorization}) into Eq.~(\ref{eq_multi_attributed_graph_matching_problem}) as follows:
\begin{equation} 
{\footnotesize
	\begin{split}
	&{F}_{gm}\left(\textbf{X}\right) = \left(\textbf{L}_{\textbf{C}}\otimes\text{vec}\left(\textbf{X}\right)\right)^{\top}\textbf{P} \left(\textbf{L}_{\textbf{C}}\otimes\text{vec}\left(\textbf{X}\right)\right) \\
	& = tr\left(\textbf{K}_{p}^{\top}\left(\textbf{L}_{\textbf{C}}^{\top}\otimes\textbf{X}\right)\right) + tr\left(\textbf{K}_{qi}^{\top}\left(\textbf{W}_{i}\otimes\textbf{Y}\right)\right) + tr\left(\textbf{K}_{qt}^{\top}\left(\textbf{W}_{t}\otimes\textbf{Z}\right)\right), \\
	&\quad\quad\quad s.t. \begin{cases}  
	\textbf{Y}=(\textbf{G}_{1i}^{\top}\textbf{X}\textbf{G}_{2i}) \circ (\textbf{H}_{1i}^{\top}\textbf{X}\textbf{H}_{2i})\\ 
	\textbf{W}_{i}= (\textbf{L}_{\textbf{C}}^{\top}\textbf{L}_{\textbf{G}i}) \circ (\textbf{L}_{\textbf{C}}^{\top}\textbf{L}_{\textbf{H}i})\\
	\textbf{Z}=(\textbf{G}_{1t}^{\top}\textbf{X}\textbf{G}_{2t}) \circ (\textbf{H}_{1t}^{\top}\textbf{X}\textbf{H}_{2t})\\
	\textbf{W}_{t}= (\textbf{L}_{\textbf{C}}^{\top}\textbf{L}_{\textbf{G}t}) \circ (\textbf{L}_{\textbf{C}}^{\top}\textbf{L}_{\textbf{H}t})
	\end{cases}
	\end{split}}
	\label{eq_F_gm}
\end{equation}
where $\mathbf{Y}$ and $\mathbf{Z}$ are the pairwise intra- and inter-layer indicator matrices, and $\circ$ is the Hadamard product. $\mathbf{K}_{qi}$ and $\mathbf{K}_{qt}$ can be divided again into $\mathbf{K}_{qi}=\mathbf{U}\mathbf{V}^{\top}$ and $\mathbf{K}_{qt}=\mathbf{S}\mathbf{T}^{\top}$ respectively by using any factorization method such as the singular value decomposition (SVD).
By using the decomposed matrices, Eq.~(\ref{eq_F_gm}) can be reformulated as follows:
\begin{equation}
{\footnotesize
	\begin{split}
	&{F}_{gm}\left(\textbf{X}\right) = tr\left(\textbf{K}_{p}^{\top}\left(\textbf{L}_{\textbf{C}}^{\top}\otimes\textbf{X}\right)\right) + \sum_{m,n}{\left[\mathbf{\Lambda}_{i}\right]_{n,n}tr\left({\mathbf{A}^{1^{\top}}_{m}}\mathbf{X}\mathbf{A}^{2}_{m,n}\mathbf{X}^{\top}\right)} \\
	&\quad\quad\quad\quad + \sum_{m,n}{\left[\mathbf{\Lambda}_{t}\right]_{n,n}tr\left({\mathbf{B}^{1^{\top}}_{m}}\mathbf{X}\mathbf{B}^{2}_{m,n}\mathbf{X}^{\top}\right)},\\
	& s.t. \begin{cases}  
	N_{qi}^{L}={N}_{L},N_{qt}^{L}={ _{{N}_{L}}\text{P}_{2}}\\
	\mathbf{\Lambda}_{i}=\left(\mathbf{L}_{\mathbf{H}i}^{\top}\mathbf{L}_{\mathbf{C}}\mathbf{L}_{\mathbf{C}}^{\top}\mathbf{L}_{\mathbf{G}i}\right), \mathbf{\Lambda}_{t}=\left(\mathbf{L}_{\mathbf{H}t}^{\top}\mathbf{L}_{\mathbf{C}}\mathbf{L}_{\mathbf{C}}^{\top}\mathbf{L}_{\mathbf{G}t}\right)\\
	\mathbf{v}_{m}={[\mathbf{v}_{m,1}^{\top}\mathbf{v}_{m,2}^{\top}\cdots\mathbf{v}_{m,N_{qi}^{L}}^{\top}]}^{\top}, \mathbf{t}_{m}={[\mathbf{t}_{m,1}^{\top}\mathbf{t}_{m,2}^{\top}\cdots\mathbf{t}_{m,N_{qt}^{L}}^{\top}]}^{\top}\\ 
	\mathbf{A}_{m}^{1}=\mathbf{G}_{1i}\text{diag}\left(\mathbf{u}_{m}\right)\mathbf{H}_{1i}^{\top}, \mathbf{A}_{m,n}^{2}=\mathbf{G}_{2i}\text{diag}\left(\mathbf{v}_{m,n}\right)\mathbf{H}_{2i}^{\top} \\
	\mathbf{B}_{m}^{1}=\mathbf{G}_{1t}\text{diag}\left(\mathbf{s}_{m}\right)\mathbf{H}_{1t}^{\top},  \mathbf{B}_{m,n}^{2}=\mathbf{G}_{2t}\text{diag}\left(\mathbf{t}_{m,n}\right)\mathbf{H}_{2t}^{\top} \\
	\end{cases}	
	\end{split}
	\label{eq_F_gm_re}}
\end{equation}
where $\mathbf{u}_m$, $\mathbf{v}_m$, $\mathbf{s}_m$, and $\mathbf{t}_m$ are ${m}^{th}$ column vectors of the factorized matrices $\mathbf{U}$, $\mathbf{V}$, $\mathbf{S}$, and $\mathbf{T}$ respectively.\footnote{A formula $tr\left({\left(\mathbf{u}{\mathbf{v}}^{\top}\right)}^{\top}\left(\mathbf{A}\circ\mathbf{B}\right)\right)=tr\left(\text{diag}\left(\mathbf{u}\right)\mathbf{A}\text{diag}\left(\mathbf{v}\right)\mathbf{B}^{\top}\right)$ is always satisfied for $\mathbf{u}\in\mathbb{R}^{m}$, $\mathbf{v}\in\mathbb{R}^{n}$ and $\mathbf{A},\mathbf{B}\in\mathbb{R}^{m\times n}$ \cite{zhou2012factorized}.}
$\mathbf{\Lambda}_{i}$ and $\mathbf{\Lambda}_{t}$ describe the confidence of each layer, and distribute the confidence information into intra- and inter-layer edges.
$\mathbf{v}_{m,n}$ and $\mathbf{t}_{m,n}$ are ${n}^{th}$ block vectors that are extracted by separating $\mathbf{v}_{m}$ and $\mathbf{t}_{m}$ according to ${N}_{qi}^{L}$ and ${N}_{qt}^{L}$, which are the numbers of sub-matrices of intra- and inter-layer affinity matrices.
$\left[\mathbf{\Lambda}_{i}\right]_{n,n}$ and $\left[\mathbf{\Lambda}_{t}\right]_{n,n}$ are ${n}^{th}$ diagonal terms of $\mathbf{\Lambda}_{i}$ and $\mathbf{\Lambda}_{t}$ respectively.

Given the reformulated objective function Eq.~(\ref{eq_F_gm_re}), we can derive convex and concave relaxations of the function similar to the single-layer factorized graph matching algorithm~\cite{zhou2012factorized,zhou2013deformable} as follows:
\begin{equation}
{\footnotesize
	\begin{split}
	&{F}_{vex}\left(\mathbf{X}\right) = {F}_{gm}\left(\mathbf{X}\right) - \frac{1}{2}{F}_{con}\left(\mathbf{X}\right)\\
	&=tr\left(\textbf{K}_{p}^{\top}\left(\textbf{L}_{\textbf{C}}^{\top}\otimes\textbf{X}\right)\right) - \frac{1}{2}\sum_{m,n}{\left[\mathbf{\Lambda}_{i}\right]_{n,n}{\left\|\mathbf{X}^{\top}\mathbf{A}_{m}^{1}-\mathbf{A}_{m,n}^{2}\mathbf{X}^{\top}\right\|}_{F}^{2} }\\
	&\quad\quad\quad-\frac{1}{2}\sum_{m,n}{\left[\mathbf{\Lambda}_{t}\right]_{n,n}{\left\|\mathbf{X}^{\top}\mathbf{B}_{m}^{1}-\mathbf{B}_{m,n}^{2}\mathbf{X}^{\top}\right\|}_{F}^{2} }
	\end{split}
	\label{eq_F_vex}}
\end{equation}
\vspace{-15pt}
\begin{equation}
{\footnotesize
	\begin{split}
	&{F}_{cav}\left(\mathbf{X}\right) = {F}_{gm}\left(\mathbf{X}\right) + \frac{1}{2}{F}_{con}\left(\mathbf{X}\right)\\
	&=tr\left(\textbf{K}_{p}^{\top}\left(\textbf{L}_{\textbf{C}}^{\top}\otimes\textbf{X}\right)\right) + \frac{1}{2}\sum_{m,n}{\left[\mathbf{\Lambda}_{i}\right]_{n,n}{\left\|\mathbf{X}^{\top}\mathbf{A}_{m}^{1}+\mathbf{A}_{m,n}^{2}\mathbf{X}^{\top}\right\|}_{F}^{2} }\\
	&\quad\quad\quad +\frac{1}{2}\sum_{m,n}{\left[\mathbf{\Lambda}_{t}\right]_{n,n}{\left\|\mathbf{X}^{\top}\mathbf{B}_{m}^{1}+\mathbf{B}_{m,n}^{2}\mathbf{X}^{\top}\right\|}_{F}^{2} }
	\end{split}
	\label{eq_F_cav}}
\end{equation}
\vspace{-15pt}
\begin{equation}
{\footnotesize
	\begin{split}
	&{F}_{con}\left(\mathbf{X}\right) =\\ 
	&\sum_{m,n}{\left[\mathbf{\Lambda}_{i}\right]_{n,n}\left(tr\left({\mathbf{A}_{m}^{1^{\top}}}\mathbf{X}\mathbf{X}^{\top}\mathbf{A}_{m}^{1}\right)+tr\left(\mathbf{A}_{m,n}^{2}\mathbf{X}^{\top}\mathbf{X}{\mathbf{A}_{m,n}^{2^{\top}}}\right)\right)}\\
	&+\sum_{m,n}{\left[\mathbf{\Lambda}_{t}\right]_{n,n} \left(tr\left({\mathbf{B}_{m}^{1^{\top}}}\mathbf{X}\mathbf{X}^{\top}\mathbf{B}_{m}^{1}\right)+tr\left(\mathbf{B}_{m,n}^{2}\mathbf{X}^{\top}\mathbf{X}{\mathbf{B}_{m,n}^{2^{\top}}}\right)\right)}\\
	\end{split}
	\label{eq_F_con}}
\end{equation}
where ${F}_{con}$ is a constant when $\mathbf{X}$ is a permutation matrix $(\eg~\mathbf{X}\mathbf{X}^{\top}=\mathbf{X}^{\top}\mathbf{X}=\mathbf{I})$.
Since the Hessian of ${F}_{vex}\left(\mathbf{X}\right)$ with respect to $\text{vec}\left(\mathbf{X}\right)$ is always negative semidefinite, ${F}_{vex}\left(\mathbf{X}\right)$ of the maximization problem is a convex function.
Similarly, since the Hessian of ${F}_{cav}\left(\mathbf{X}\right)$ is always positive semidefinite, ${F}_{cav}\left(\mathbf{X}\right)$ is a concave function.
However, since $\mathcal{G}_{1}$ and $\mathcal{G}_{2}$ often have different sizes in practical application, $\mathbf{X}$ cannot satisfy the condition as a permutation matrix in general.
To convert $\mathbf{X}$ to a permutation matrix, we add dummy nodes to the smaller graph as presented in \cite{zhou2012factorized,zhou2013deformable}.
Details of the proof are presented in the supplementary materials.

\subsection{Multi-layer path following algorithm}
\label{sec_multi_layer_path_following}

\begin{algorithm}[t!]
	\small
	\renewcommand{\algorithmicrequire}{\textbf{Input:}}
	\renewcommand{\algorithmicensure}{\textbf{Output:}}
	\caption{Multi-layer Factorized Graph Matching}
	\label{alg_matching_framework}      
	\begin{algorithmic}[1]   
		\small                 
		\REQUIRE Affinity matrices \{$\mathbf{K}_{p}$, $\mathbf{K}_{qi}$, $\mathbf{K}_{qt}$\}, Edge incidence matrices \{$\mathbf{G}_{1i}$, $\mathbf{G}_{2i}$, $\mathbf{G}_{1t}$, $\mathbf{G}_{2t}$, $\mathbf{H}_{1i}$, $\mathbf{H}_{2i}$, $\mathbf{H}_{1t}$, $\mathbf{H}_{2t}$\}, Layer incidence matrices \{$\mathbf{L}_{\mathbf{G}i}$, $\mathbf{L}_{\mathbf{G}t}$, $\mathbf{L}_{\mathbf{H}i}$, $\mathbf{L}_{\mathbf{H}t}$\}
		\ENSURE Assignment matrix $\hat{\mathbf{X}}$		
		\STATE Initialize $\mathbf{X}$ to be a double stochastic matrix
		\STATE Initialize $\mathbf{L}_{\mathbf{C}}$ to be a uniform vector
		\STATE Factorize $\mathbf{K}_{qi}$ and $\mathbf{K}_{qt}$ using SVD
		\STATE \textbf{for} $\theta = 0$ to $1$
		\STATE \quad Solve $\hat{\textbf{X}} = {arg\max }_{ \textbf{X} } {F}_{\theta}\left(\textbf{X}\right)$
		\STATE \quad Update $\mathbf{X}\leftarrow\hat{\mathbf{X}}$
		\STATE \quad Update $\mathbf{L}_{\mathbf{C}}\leftarrow\text{LayerConfidence}\left(\mathbf{X},\mathbf{K}_{qi}\right)$
		\STATE \quad Update $\mathbf{W}_{i}$, $\mathbf{W}_{t}$, $\mathbf{\Lambda}_{i}$ and $\mathbf{\Lambda}_{t}$ using $\mathbf{L}_{\mathbf{C}}$
		\STATE \textbf{end}
	\end{algorithmic}
\end{algorithm}

The objective function of the proposed multi-layer path following algorithm can be constructed by combining two relaxed functions (Eq.~(\ref{eq_F_vex}) and Eq.~(\ref{eq_F_cav})) as follows:
\begin{equation}
\begin{split}
{F}_{\theta}\left(\mathbf{X}\right) = \left(1-\theta\right){F}_{vex}\left(\mathbf{X}\right) + \theta{F}_{cav}\left(\mathbf{X}\right)\\
\end{split}
\label{eq_F_alpha}
\end{equation}
where $\theta$ is a control parameter that is gradually increased during the optimization process.
We fixed the step size for increasing $\theta$ to 0.01 in all experiments.
Based on this objective function, the multi-layer path following algorithm can be designed as shown in Algorithm~\ref{alg_matching_framework}.
Since ${F}_{\theta}\left(\mathbf{X}\right)$ is a convex function at the start of the algorithm, we can find a global optimum by using convex optimization methods such as modified Frank-Wolfe (FW) algorithm~\cite{fukushima1984modified}.
During the series of iteration steps, ${F}_{\theta}\left(\mathbf{X}\right)$ is gradually changed to a concave function that has the same global optimum as the original objective function.
At the same time, the optimum of ${F}_{\theta}\left(\mathbf{X}\right)$ also approaches the globally optimal solution.

In the middle of iteration steps (Lines 7-8), the path is modified based on the proposed layer confidence measure that evaluates relative confidences among attributes. 
This modification encourages the optimization algorithm to follow the information of more confident attributes.
Details of the measure are presented in Sec.~\ref{sec_relative_confidence_computation}.

\subsection{Relative confidence computation}
\label{sec_relative_confidence_computation}

To define a layer confidence measure, we first assume that true correspondences have strong affinities with themselves, while false correspondences have relatively weak connections with others; because the correctly matched graphs should have similar structures. 
This assumption has been used frequently in recent works that follow the spectral matching scheme~\cite{cho2010reweighted,cho2014finding,cour2006balanced,leordeanu2005spectral}.
Based on this, we use the difference between the connection strength of true/false clusters as a confidence measure because the difference is decreased when true/false clusters become jumbled.

The connection strength can be determined by computing the average affinity value of each cluster.
However, since true correspondences are unidentified during the matching process, we temporally use the current solution as a true/false indicator matrix.
Finally, the proposed layer confidence measure is defined as follows:
%
\begin{equation}
{\small		
	\begin{split}
	&\text{LayerConfidence}\left(\mathbf{X},\mathbf{K}_{qi}\right) = \mathbf{L}_{\mathbf{C}} = \mathbf{L}_{\mathbf{C}true} - \mathbf{L}_{\mathbf{C}false},\\
	&\left[\mathbf{L}_{\mathbf{C}true}\right]_{\alpha}= \text{mean}\left(\text{vec}(\mathbf{K}^{\alpha;\alpha}_{qi}\circ(\mathbf{G}^{\top}_{1i}\mathbf{X}\mathbf{G}_{2i})\circ(\mathbf{H}^{\top}_{1i}\mathbf{X}\mathbf{H}_{2i}))\right), \\
	&\left[\mathbf{L}_{\mathbf{C}false}\right]_{\alpha}= \text{mean}\left(\text{vec}(\mathbf{K}^{\alpha;\alpha}_{qi}\circ(\mathbf{G}^{\top}_{1i}\mathbf{\overline{X}}\mathbf{G}_{2i})\circ(\mathbf{H}^{\top}_{1i}\mathbf{\overline{X}}\mathbf{H}_{2i}))\right),
	\end{split}
	\label{eq_layer_confidence}}
\end{equation}
where $\mathbf{X}$ is a discretized current solution by using discretization methods such as the Hungarian method~\cite{munkres1957algorithms}, and $\mathbf{\overline{X}}$ is the binary complement vector of $\mathbf{X}$.

\subsection{Discussion: Memory efficiency}
\label{sec_discussion_scalability}
As stated above, the multi-layer structure has a scalability problem caused by the huge size of the supra-adjacency matrix.
The space requirements grow as $O\left({N}_{v}^{4}{N}_{L}^{2}\right)$, where ${N}_{v}$ is the number of vertices and ${N}_{L}$ is the number of layers.
On the other hand, the proposed algorithm has fewer space requirements $O\left({N}_{v}^{4}{N}_{L}+{N}_{v}^{2}{N}_{L}^{2}\right)$.
Since the number of attributes greatly affects the expressivity and robustness of the algorithm, the proposed factorization relieves the scalability problem when we increase the number of attributes. 
Moreover, because most matrix operations can be performed before starting the optimization process, the proposed algorithm can take advantage of the computational complexity aspect.

\section{Experimental Results}
\label{sec_experimental_results}

To evaluate the proposed algorithm, we design two experiments using the synthetic dataset and WILLOW dataset~\cite{cho2013learning}.\footnote{http://www.di.ens.fr/willow/research/graphlearning/}
We compare our algorithm with the well-known graph matching methods such as graduated assignment graph matching (GAGM)~\cite{gold1996graduated}, spectral matching (SM)~\cite{leordeanu2005spectral}, spectral matching with affine constraints (SMAC)~\cite{cour2006balanced}, dual decomposition approach (DD)~\cite{torresani2008feature,torresani2013dual}, integer projected fixed points matching (IPFP)~\cite{leordeanu2009integer}, reweighted random walk matching (RRWM)~\cite{cho2010reweighted}, factorized graph matching (FGM)~\cite{zhou2012factorized}, max pooling matching (MPM)~\cite{cho2014finding}, graduated nonconvexity and concavity procedure (GNCCP)~\cite{liu2014gnccp}, Hungarian belief propagation (HBPM)~\cite{zhang2016pairwise}, and multi-layer random walk matching (MLRWM)~\cite{park2016multi}.
The evaluation framework is implemented based on the open MATLAB programs of \cite{cho2010reweighted} and \cite{park2016multi}.
The compared methods are also attached from the authors' open-source code, and their parameters are selected as those given in the original papers. 

\subsection{Performance evaluation for synthetic dataset}
\label{sec_experimental_results_synthetic}

We evaluate the proposed algorithm for synthetic graph matching problems.
To generate a pair of synthetic graphs, we use the same experimental scheme outlined in \cite{park2016multi}.
First, an initial graph $\mathcal{G}_{0}$ is generated, which has several types of attributes $\mathcal{A}_{0}$.
To assign different characteristics into each attribute, we randomly generate attribute values with different variances for each layer.
Then, we generate two graphs $\mathcal{G}_{1}$ and $\mathcal{G}_{2}$ by adding small attribute deformations that follow a Gaussian distribution $\mathcal{N}\left(0,{\epsilon}^{2}\right)$ and the same number of randomly defined outliers ${v}_{out}$ into $\mathcal{G}_{0}$.
Each element of an intra-layer affinity matrix $\left[\mathbf{K}^{\alpha;\alpha}\right]_{ia,jb}$ is defined as follows:
\begin{equation}
\left[\mathbf{K}^{\alpha;\alpha}\right]_{ia,jb} = \text{exp}(-{|(1-\omega) + \omega(\left[{r}^{\alpha}_{1}\right]_{ij}-\left[{r}^{\alpha}_{2}\right]_{ab})|}_{2}/{\sigma}^{2}),
\end{equation} 
where $\left[{r}^{\alpha}_{1}\right]_{ij}$ and $\left[{r}^{\alpha}_{2}\right]_{ab}$ are randomly assigned edge attributes of layer $\alpha$, and $\sigma$ is a scaling factor that is fixed as $0.3$ in these experiments.
$\omega$ is a control parameter that determines the variance of the affinity values and reflects different characteristics of each attribute.
$\omega$ is randomly selected in $[0.1,1]$.
To integrate the attributes for single-layer graph matching algorithms, we normalize the affinity matrices individually and aggregate the matrices by summing them.

\begin{figure*}[t!]
	\begin{tabular}{c c c}
		\includegraphics[width=0.32\linewidth]{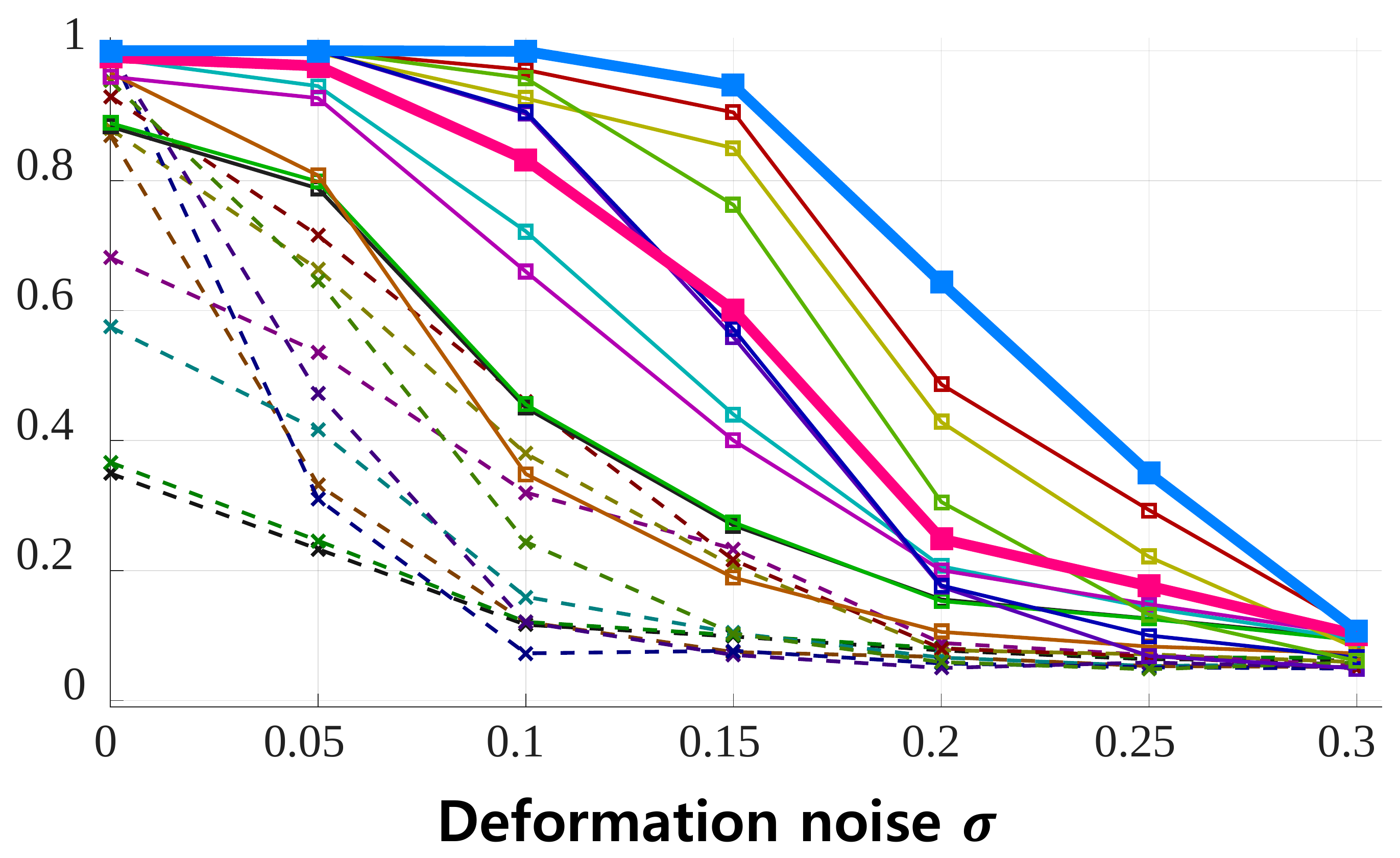}&\includegraphics[width=0.32\linewidth]{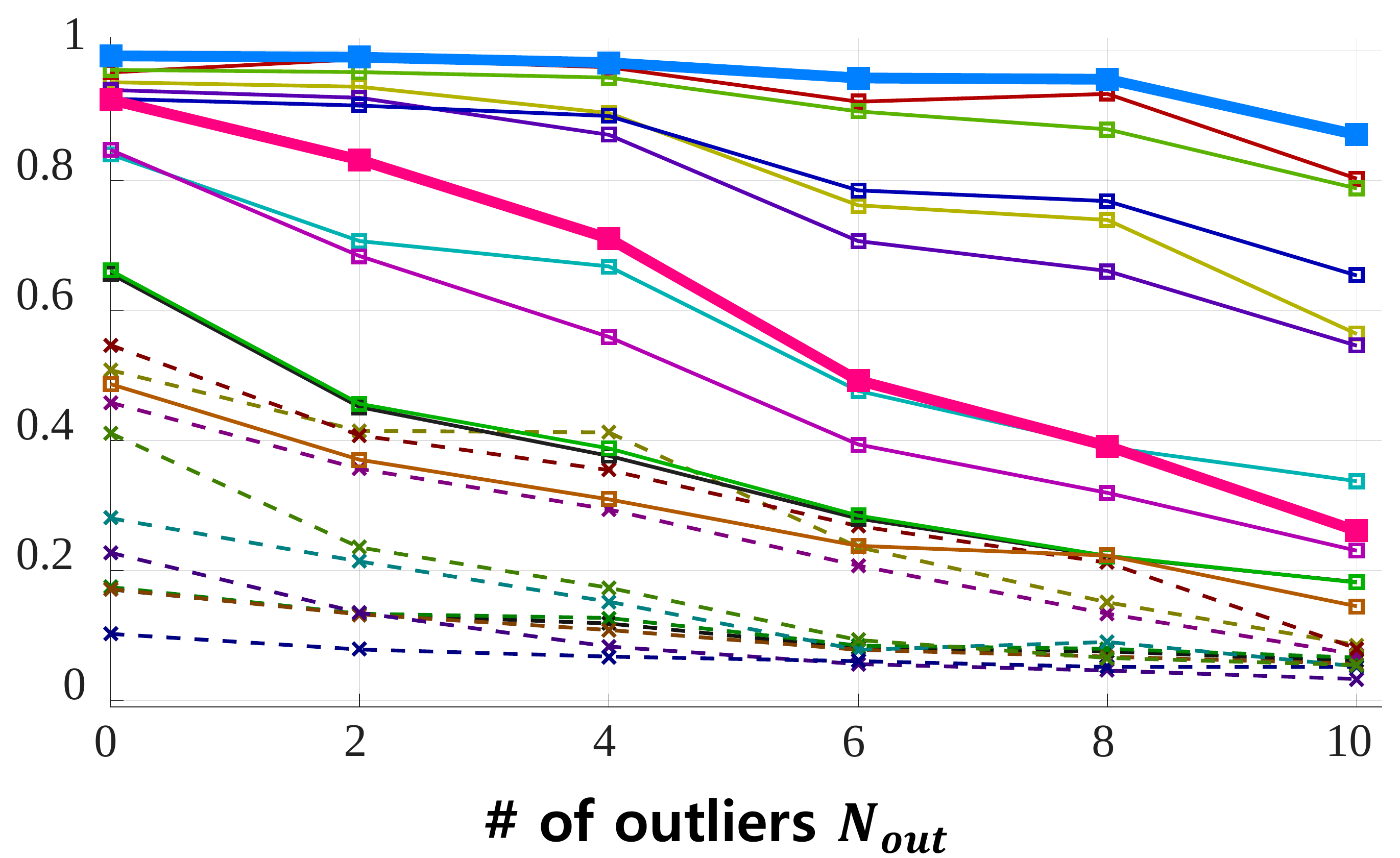}&\includegraphics[width=0.32\linewidth]{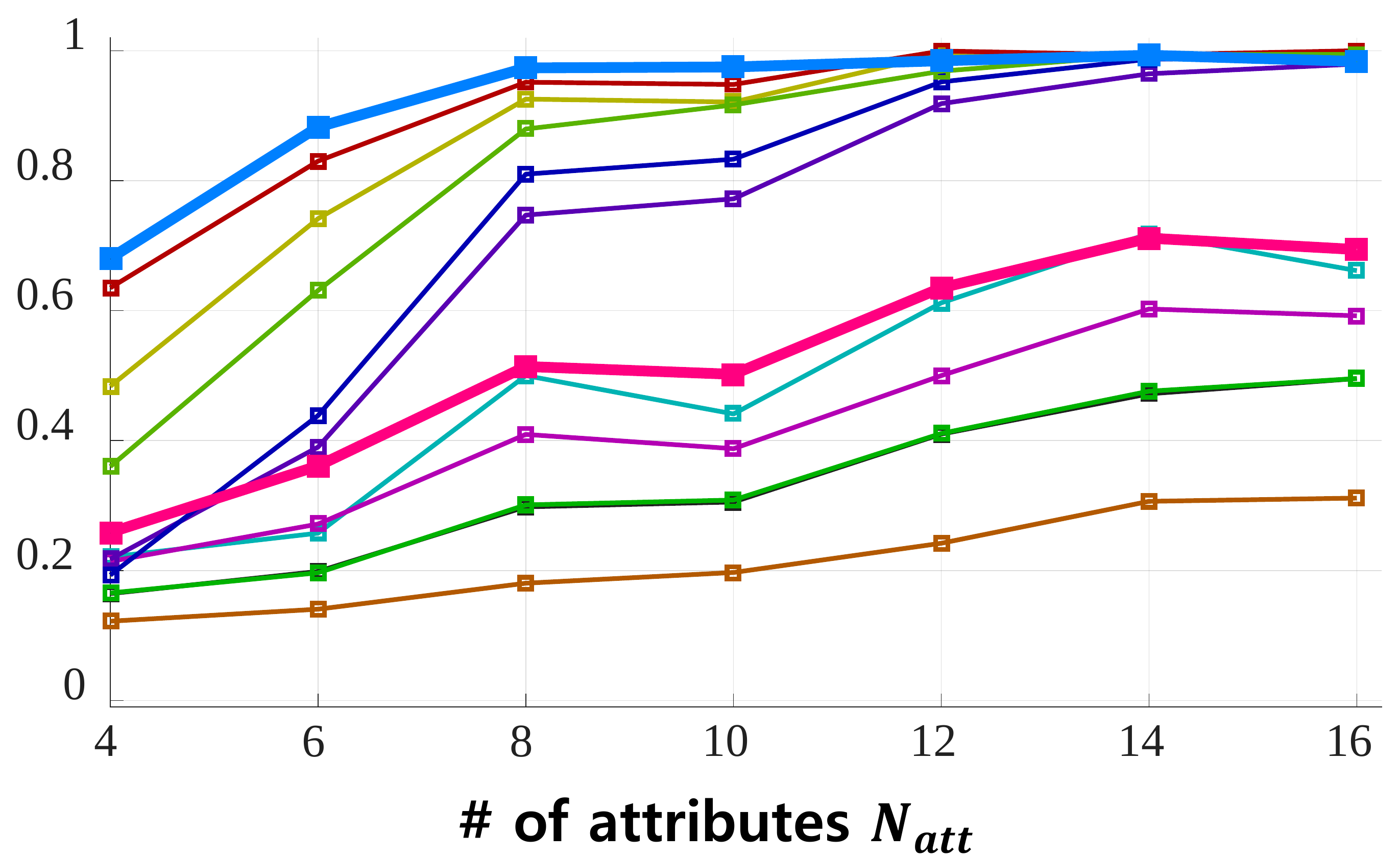}\\
		\includegraphics[width=0.32\linewidth]{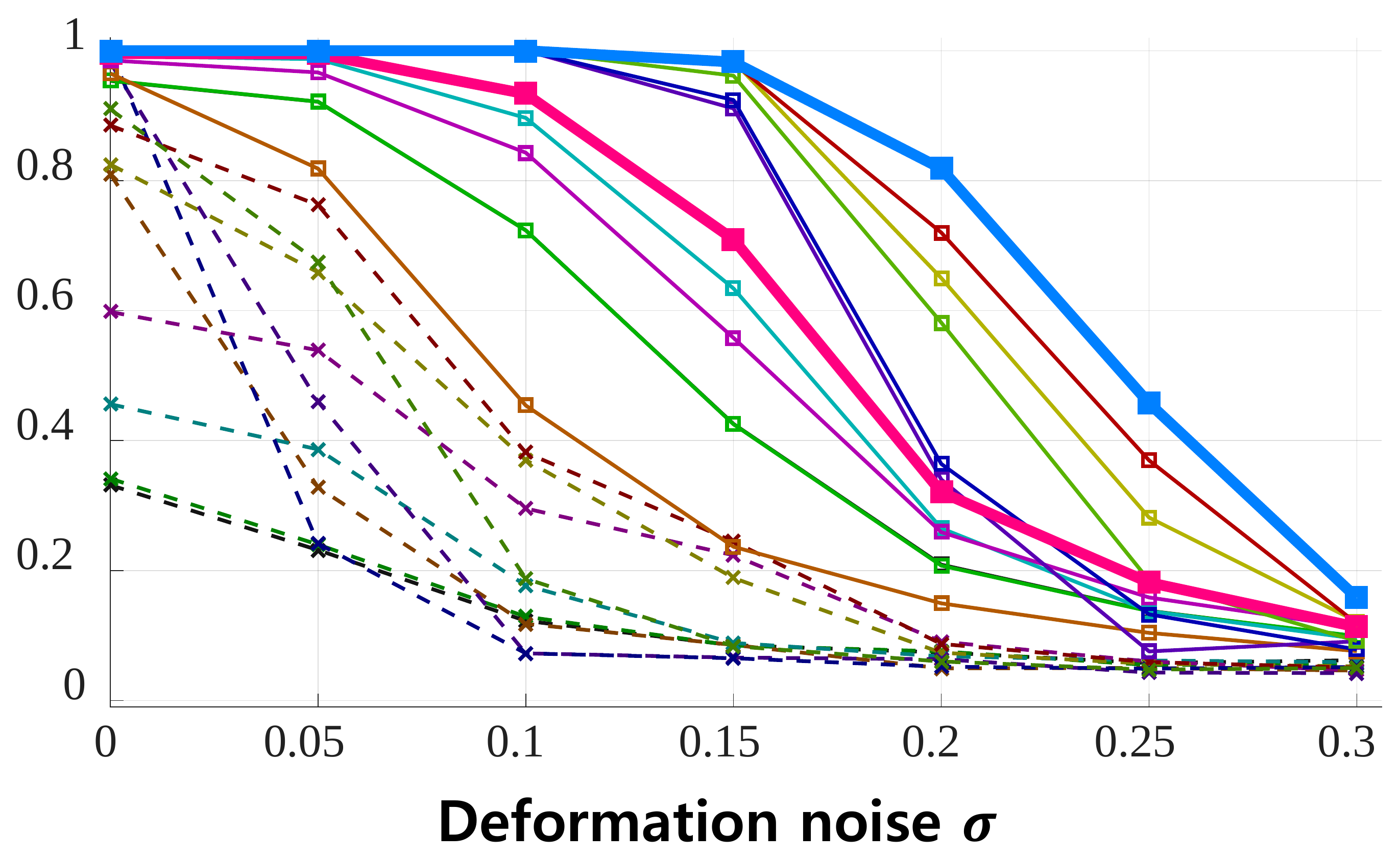}&\includegraphics[width=0.32\linewidth]{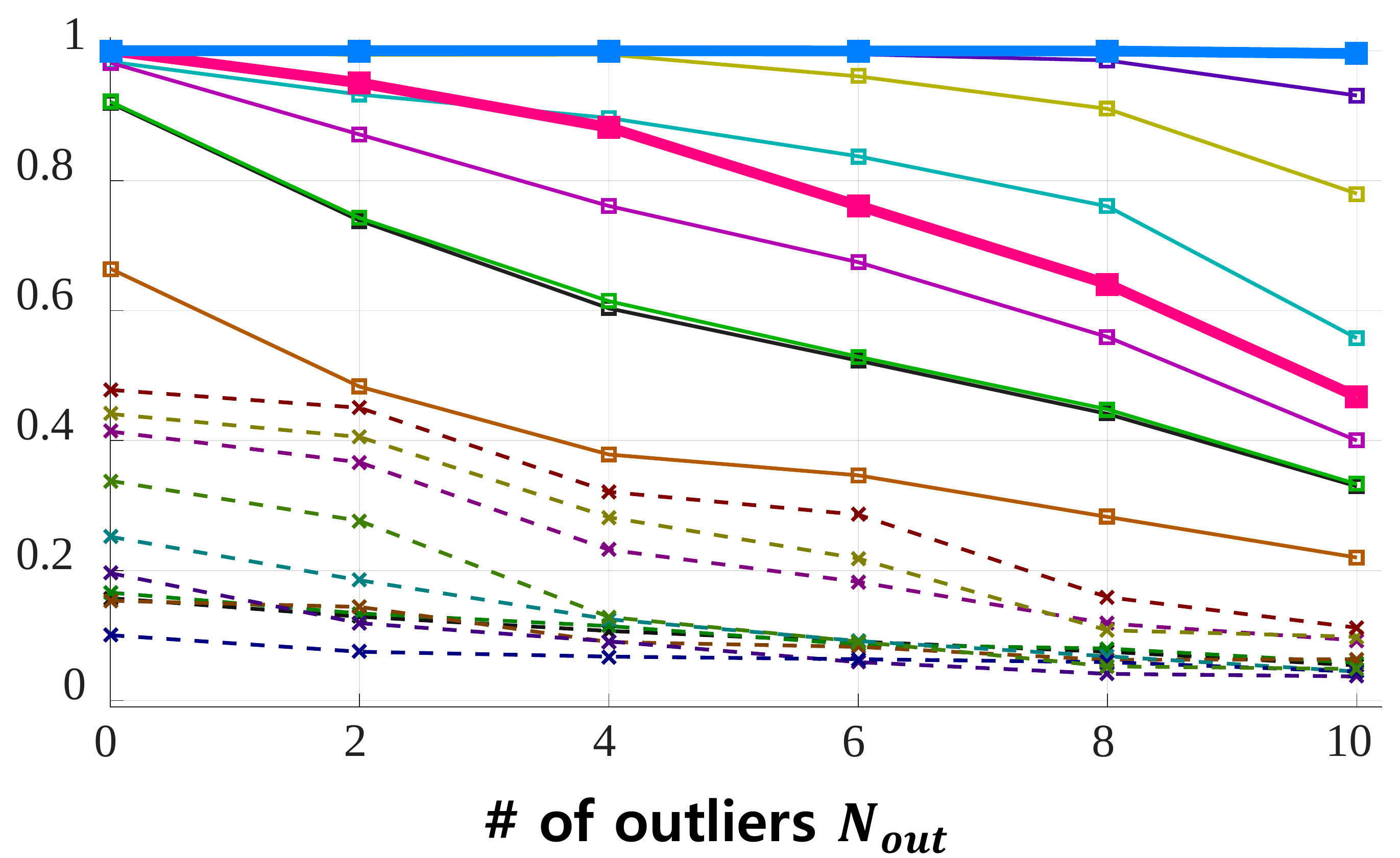}&\includegraphics[width=0.32\linewidth]{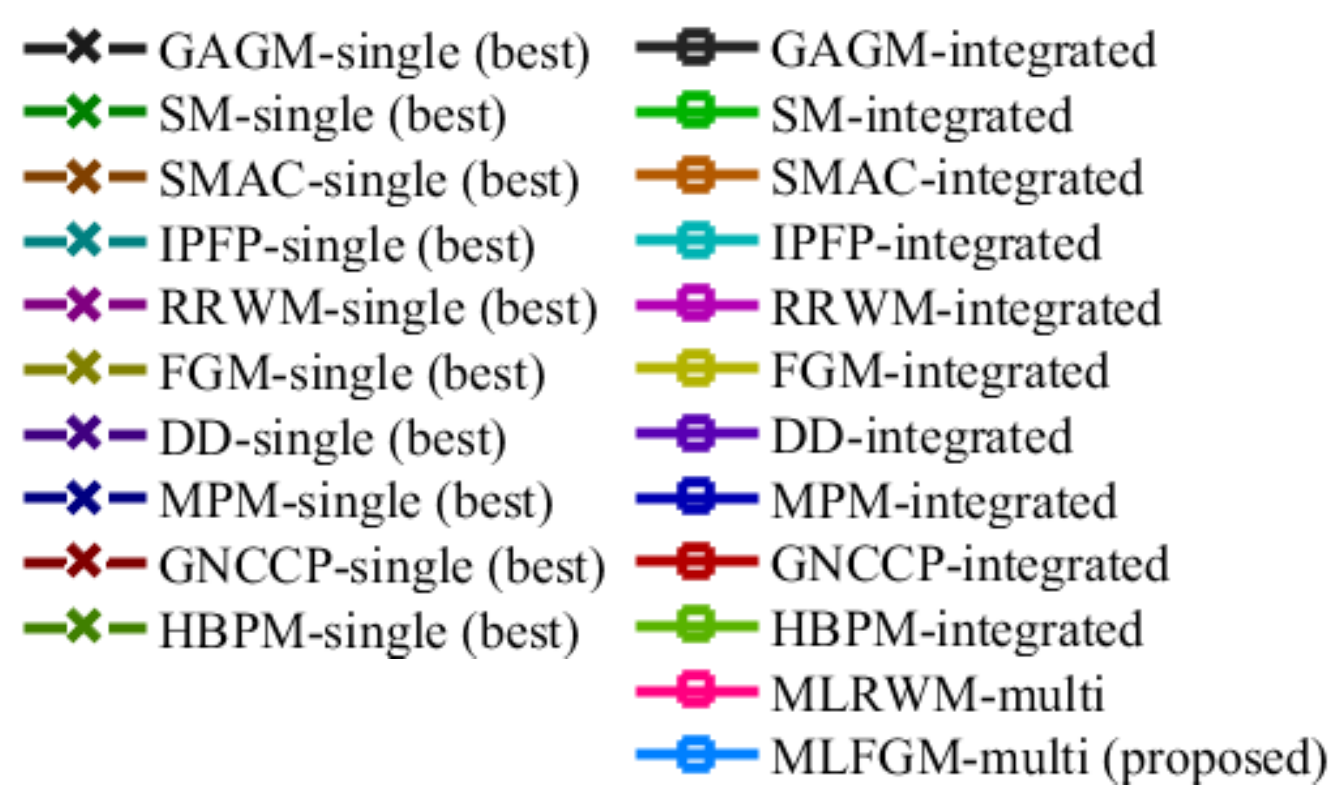}
	\end{tabular} \vspace{-8pt}
	\caption{ Synthetic graph matching results. Columns: each type of experiments -- deformation, outlier, and attributes. Rows: number of attributes -- five and ten attributes.}
	\label{fig_experiments_synthetic} \vspace{-4pt}
\end{figure*}

\begin{table*}[t]
	\centering
	\caption{Parameter setting for synthetic graph matching experiments} 
	\label{tab_param_synthetic}
	{ 
		\vspace{-8pt}
		\small 
		\begin{tabular}{l|c|c}
			\noalign{\hrule height 0.5pt} \hline			
			Experiments      & Varied parameter                    & Fixed parameters                                    \\ \hline		\hline
			Deformation      &    $\epsilon=0$ -- $0.3$     &  ${N}_{att}=5,10$, ${N}_{in}=20$, ${N}_{out}=2$, ${\sigma}^{2}=0.3$   \\
			Outlier          &    ${N}_{out}=0$ -- $10$     &  ${N}_{att}=5,10$, ${N}_{in}=20$, $\epsilon=0.1$, ${\sigma}^{2}=0.3$  \\
			Attributes       &    ${N}_{att}=4$ -- $16$     &  ${N}_{in}=20$, ${N}_{out}=4$, $\epsilon=0.15$, ${\sigma}^{2}=0.3$      \\ 
			\noalign{\hrule height 0.5pt} \hline	
		\end{tabular} 
	} \vspace{-10pt}
\end{table*}

Then, we performed three types of experiments: deformation, outlier, and attributes, as listed in Table~\ref{tab_param_synthetic}.
In the deformation experiment, each pair of graphs is generated with different magnitudes of deformation $\epsilon$, while the other parameters are fixed.
By contrast, in the outlier experiment, only the number of outliers ${N}_{out}$ is varied.
In the last experiment, both of the parameters are fixed, and the number of attributes is varied.
For all experiments, the number of inliers is fixed as 20, and each experiment is iteratively performed 100 times.

As shown in Fig.~\ref{fig_experiments_synthetic}, the proposed algorithm (`MLFGM-multi') exhibits better performance than most of the other methods.\footnote{The postfix `-single', `-integrated', and `-multi' indicate the results obtained using a single attribute, an integrated  attribute, and multiple attributes, respectively. Note that we only present the best result among all the `-single' results for each method for comparison.}
Moreover, as stated in \cite{park2016multi}, this synthetic graph matching scheme cannot fully exploit practical environments that contain real attributes having complex properties.
Thus, to evaluate our algorithm in realistic environments, we perform the second experiment using general real images in Sec.~\ref{sec_experimental_results_willow}.



\subsection{Performance evaluation for WILLOW dataset}
\label{sec_experimental_results_willow}

In this experiment, we use graphs that are constructed from real images in the WILLOW dataset~\cite{cho2013learning} to evaluate the proposed algorithm in realistic environments.
The WILLOW dataset consists of five classes: \textit{face}, \textit{motorbike}, \textit{car}, \textit{duck}, and \textit{winebottle}, and provides annotations that reflect each object structure.
To generate a graph from each image, we first extract interest points by using the Hessian detector~\cite{murphy2012machine}. 
Then, we select the 10 nearest interest points from the annotated points as inliers.
To consider variations caused by deformation and outliers, we randomly select pairs of images for each category, and include outliers that are randomly selected among the remaining interest points.
The number of outliers is varied from 0 to 10 for each category, and each test is iterated 100 times for each category.

\begin{table}[t]
	\centering
	\caption{Attribute description} \vspace{-8pt}
	\label{tab_attribute_list}
	{
		\small
        \renewcommand{\tabcolsep}{1.5mm}
		\begin{tabular}{m{1.4cm}|m{0.8cm}|m{5.15cm}}
			\noalign{\hrule height 0.5pt} \hline
			Types						&	Abbr.			& Description			     				\\ \hline	\hline
			\multirow{2}{*}{Appearance}	&	CSiD       		& \textit{Concatenated} SIFT descriptor difference  	\\ 
			&	CCoD			& \textit{Concatenated} color histogram difference	\\  \hline
			\multirow{2}{*}{Geometric}	&	RDHD			& Relative \textit{distance} histogram difference	\\ 
			&	RAHD			& Relative \textit{angle} histogram difference		\\ \hline
			Multi						&	Multi			& Combination of all attributes    \\ 
			\noalign{\hrule height 0.5pt} \hline
		\end{tabular}
	} \vspace{-13pt}
\end{table}

We define four types of attributes as listed in Table~\ref{tab_attribute_list}.
CSiD and CCoD are appearance attributes that describe each edge attribute by concatenating the SIFT descriptors~\cite{lowe2004distinctive} of two interest points and RGB color histograms respectively.
The geometric attributes, RDHD and RAHD, are generated by separating the log-polar histogram of HARG~\cite{cho2013learning} into relative distance and angle histograms.
Then, we use the normalized Hamming distance~\cite{cho2013learning} to compute differences between two attribute vectors, which are defined as affinity values.
The affinity matrices are normalized again for each layer to adjust the scales among attributes.
Finally, an integrated affinity matrix is constructed by summing the normalized matrices.

\begin{figure*}[bt!]
	\centering
	\begin{subfigure}{0.92\linewidth}
		\includegraphics[width=\linewidth]{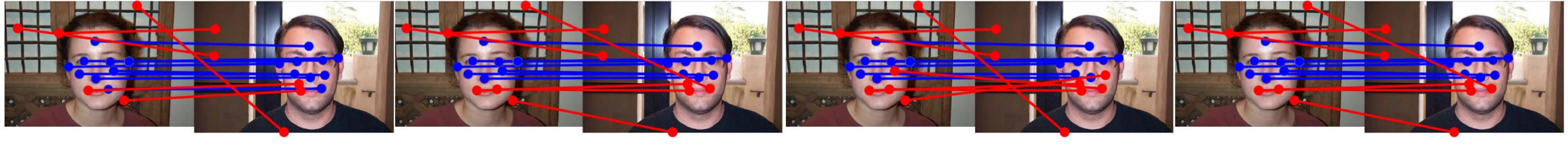}
		\caption{WILLOW-\textit{face}: MLFGM(9/10), MLRWM(8/10), FGM(7/10), RRWM(8/10)}
		\label{fig_Result_image_face}		
	\end{subfigure}			
	\begin{subfigure}{0.92\linewidth}		
		\includegraphics[width=\linewidth]{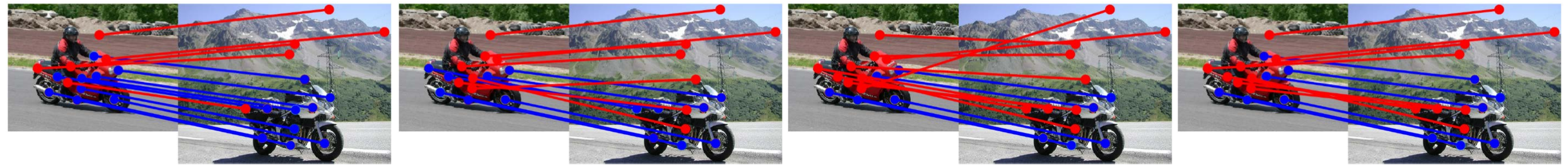}
		\caption{WILLOW-\textit{motorbike}: MLFGM(9/10), MLRWM(7/10), FGM(5/10), RRWM(6/10)}
		\label{fig_Result_image_motorbike}
	\end{subfigure}		
	\begin{subfigure}{0.92\linewidth}		
		\includegraphics[width=\linewidth]{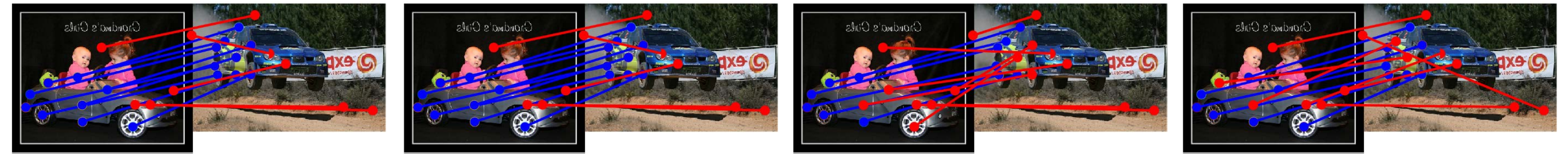}
		\caption{WILLOW-\textit{car}: MLFGM(9/10), MLRWM(9/10), FGM(6/10), RRWM(7/10)}
		\label{fig_Result_image_car}
	\end{subfigure}		
	\begin{subfigure}{0.92\linewidth}		
		\includegraphics[width=\linewidth]{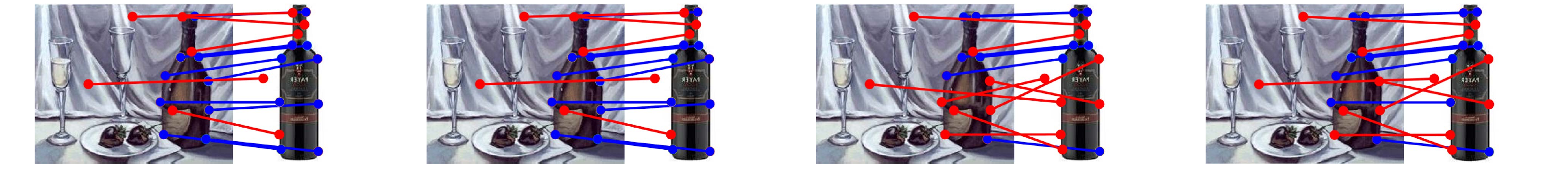}
		\caption{WILLOW-\textit{winebottle}: MLFGM(9/10), MLRWM(9/10), FGM(6/10), RRWM(7/10)}
		\label{fig_Result_image_winebottle}
	\end{subfigure}				
	\vspace{-5pt}
	\caption{
		Results for the WILLOW dataset.
		\textcolor{blue}{Blue}/\textcolor{red}{red} lines indicate \textcolor{blue}{true}/\textcolor{red}{false} correspondences respectively.
		From top to bottom: each of categories.
		From left to right: each of graph matching algorithms--MLFGM~(proposed), MLRWM~\cite{park2016multi}, FGM~\cite{zhou2012factorized}, RRWM~\cite{cho2010reweighted}. The numbers in brackets represent the numbers of correct matches and ground truth matches.       
	} 
	\vspace{-10pt}
	\label{fig_experiments_WILLOW}
\end{figure*}

As shown in Table~\ref{tab_results_WILLOW} and Fig.~\ref{fig_experiments_WILLOW}, the proposed algorithm exhibits state-of-the-art performance in all categories except the \textit{face} category --- our algorithm shows a relatively lower matching accuracy in the \textit{face} category.
This is because the proposed layer confidence measure fails when evaluating the confidence values more frequently than the other categories. 
Nevertheless, the proposed algorithm shows comparable performance to other algorithms in the category.
We expect that the problem can be resolved by improving the measure using more sophisticated learning-based methods. 
We will address this issue in future work.

\begin{table*}[tb]
	\centering
	\caption{Performance evaluation for the WILLOW dataset~\cite{cho2013learning}. \textcolor{red}{\textbf{Red}} and \textcolor{blue}{\textbf{blue}} bold numbers denote the best and the second-best performance in each category.} \vspace{-8pt}
	\label{tab_results_WILLOW}
	\scriptsize
    \renewcommand{\tabcolsep}{1.6mm}
	\begin{tabular}{l||c|c|c|c|c||c|c|c|c|c||c|c|c|c|c}
		\noalign{\hrule height 0.5pt} \hline	
		\multirow{2}{*}{} & \multicolumn{5}{c||}{WILLOW-\textbf{\textit{face}}}            & \multicolumn{5}{c||}{WILLOW-\textbf{\textit{motorbike}}} & \multicolumn{5}{c}{WILLOW-\textbf{\textit{car}}}      \\ \cline{2-16} 
		& CSiD  & CCoD & RDHD  & RAHD  & \textbf{Multi} & CSiD  & CCoD & RDHD  & RAHD  & \textbf{Multi} & CSiD  & CCoD & RDHD  & RAHD  & \textbf{Multi} \\ \noalign{\hrule height 0.5pt} \hline			
		GAGM~\cite{gold1996graduated}		&56.53& 9.63&16.55&53.55&67.02&17.15& 7.55&12.27&52.08&43.58&20.68& 7.32&12.98&49.70&44.73\\ \hline
		SM~\cite{leordeanu2005spectral}		&56.52& 9.42&16.75&54.18&66.85&17.25& 7.48&12.52&52.47&43.50&20.67& 7.12&12.88&50.05&44.82\\ \hline
		SMAC~\cite{cour2006balanced}		&52.68& 9.08&14.93&51.93&61.55&15.12& 7.37&11.72&46.70&31.90&18.75& 6.97&12.43&40.90&31.87\\ \hline
		DD~\cite{torresani2013dual}			&56.68& 9.43&23.88&58.83&\color[HTML]{3531FF} \textbf{76.60}&17.22& 7.57&14.67&53.63&52.80&20.53& 7.28&16.10&50.73&50.98\\ \hline 
		IPFP~\cite{leordeanu2009integer}	&56.63& 9.38&17.48&57.43&67.72&17.07& 7.62&13.13&53.33&43.18&20.63& 7.30&14.28&50.28&44.23\\ \hline
		RRWM~\cite{cho2010reweighted}		&56.55& 9.52&21.32&57.38&74.65&17.28& 7.38&14.78&53.65&51.55&20.58& 7.22&15.77&50.12&50.17\\ \hline
		FGM~\cite{zhou2012factorized}		&57.17& 9.32&21.07&58.73&76.23&17.13& 7.65&14.30&45.37&39.28&20.60& 6.78&14.40&42.03&35.98\\ \hline
		MPM~\cite{cho2014finding}			&57.17& 9.30&17.68&60.62&\color[HTML]{FE0000} \textbf{80.78}&16.90& 7.53&12.10&53.32&53.73&20.60& 7.58&12.97&49.18&50.22\\ \hline
		GNCCP~\cite{liu2014gnccp}			&56.77& 9.43&21.48&58.38&75.50&17.12& 7.63&14.83&53.80&52.18&20.40& 7.13&15.73&51.03&50.53\\ \hline 
		HBPM~\cite{zhang2016pairwise}		&56.77& 9.45&22.00&59.07&76.38&17.18& 7.37&15.00&53.60&52.63&20.48& 7.27&15.37&50.85&50.75\\ \noalign{\hrule height 0.5pt} \hline
		MLRWM~\cite{park2016multi}			&  -  &  -  &  -  &  -  &71.03&  -  &  -  &  -  &  -  &\color[HTML]{3531FF} \textbf{53.97}&  -  &  -  &  -  &  -  &\color[HTML]{3531FF} \textbf{51.33}\\ \hline
		MLFGM~(proposed)							&  -  &  -  &  -  &  -  &75.75&  -  &  -  &  -  &  -  &\color[HTML]{FE0000} \textbf{54.67}&  -  &  -  &  -  &  -  &\color[HTML]{FE0000} \textbf{51.95}\\ \hline

		\multirow{2}{*}{} & \multicolumn{5}{c||}{WILLOW-\textbf{\textit{duck}}}             & \multicolumn{5}{c||}{WILLOW-\textbf{\textit{winebottle}}} & \multicolumn{5}{c}{\textbf{Average}}             \\ \cline{2-16} 
		& CSiD  & CCoD & RDHD  & RAHD  & \textbf{Multi} & CSiD  & CCoD & RDHD  & RAHD  & \textbf{Multi} & CSiD  & CCoD & RDHD  & RAHD  & \textbf{Multi} \\ \noalign{\hrule height 0.5pt} \hline			
		GAGM~\cite{gold1996graduated}		&18.28& 7.10&12.87&44.67&39.42&24.87& 7.07&14.08&54.82&46.70&27.50& 7.73&13.75&50.96&48.29\\ \hline
		SM~\cite{leordeanu2005spectral}		&18.32& 7.07&12.83&44.32&39.87&25.33& 7.07&14.50&54.27&47.03&27.62& 7.63&13.90&51.06&48.41\\ \hline
		SMAC~\cite{cour2006balanced}		&16.83& 6.95&11.43&36.32&27.68&23.25& 7.63&12.42&46.00&34.70&25.33& 7.60&12.59&44.37&37.54\\ \hline		
		DD~\cite{torresani2013dual}			&18.40& 7.22&15.13&47.00&47.15&25.37& 7.32&17.65&55.88&56.08&27.64& 7.76&17.49&53.22&56.72\\ \hline
		IPFP~\cite{leordeanu2009integer}	&18.20& 7.25&12.43&45.98&39.43&25.18& 7.20&15.77&55.17&47.47&27.54& 7.75&14.62&52.44&48.41\\ \hline
		RRWM~\cite{cho2010reweighted}		&18.37& 7.30&14.33&46.35&46.83&25.42& 7.07&17.62&54.68&54.65&27.64& 7.70&16.76&52.44&55.57\\ \hline
		FGM~\cite{zhou2012factorized}		&18.20& 7.05&14.82&36.87&36.47&25.17& 7.28&19.25&49.03&47.20&27.65& 7.62&16.77&46.41&47.03\\ \hline
		MPM~\cite{cho2014finding}			&18.18& 7.45&11.02&44.10&46.58&25.65& 7.62&14.63&55.77&56.32&27.70& 7.90&13.68&52.60&\color[HTML]{3531FF} \textbf{57.53}\\ \hline
		GNCCP~\cite{liu2014gnccp}			&18.30& 7.08&15.07&47.05&46.85&25.33& 7.28&17.63&56.28&55.50&27.58& 7.71&16.95&53.31&56.11\\ \hline
		HBPM~\cite{zhang2016pairwise}		&18.37& 7.23&14.03&46.78&47.10&25.42& 7.25&17.92&55.95&55.67&27.64& 7.71&16.86&53.25&56.51\\ \noalign{\hrule height 0.5pt} \hline
		MLRWM~\cite{park2016multi}			&  -  &  -  &  -  &  -  &\color[HTML]{3531FF} \textbf{47.67}&  -  &  -  &  -  &  -  &\color[HTML]{3531FF} \textbf{56.83}&  -  &  -  &  -  &  -  &56.17\\ \hline	
		MLFGM~(proposed)							&  -  &  -  &  -  &  -  &\color[HTML]{FE0000} \textbf{48.27}&  -  &  -  &  -  &  -  &\color[HTML]{FE0000} \textbf{57.62}&  -  &  -  &  -  &  -  &\color[HTML]{FE0000} \textbf{57.65}\\ \noalign{\hrule height 0.5pt}
		
	\end{tabular}
\end{table*}

\section{Conclusion}
\label{sec_conclusion}

In this paper, we proposed a multi-layer factorized graph matching algorithm to solve multi-attributed graph matching problems.
First, we proposed a multi-layer graph factorization method that divides a multi-layer association graph into several incidence matrices and affinity matrices.
Then, we proposed a multi-attributed graph matching algorithm based on the path following scheme that has exhibited state-of-the-art performances for single-layer graph matching problems.
To apply the scheme, we proposed convex and concave relaxations of the original objective function by using the factorized matrices.
To evaluate the proposed method, we performed two experiments on synthetic datasets and the WILLOW dataset.
Extensive experiments demonstrated the superiority of the proposed algorithm over other state-of-the-art methods in terms of performance.

\clearpage
{\small
	\bibliographystyle{ieee}
	\bibliography{egbib}
}

\end{document}